\documentclass[10pt, a4paper]{article}
\usepackage{lrec}
\usepackage{graphicx}
\usepackage{epstopdf}
\usepackage[latin1]{inputenc}
\usepackage{hyperref}
\usepackage{booktabs, multirow}
\usepackage{caption}
\usepackage{subcaption}
\usepackage{xcolor}

\graphicspath{{./images/}}

\title{Model-based annotation of coreference}

\name{Rahul Aralikatte 
\textnormal{and} Anders S{\o}gaard
}

\address{
University of Copenhagen
\\
         \{rahul, soegaard\}@di.ku.dk\\}

\abstract{
Humans do not make inferences over texts, but over {\em models} of what texts are about. When annotators are asked to annotate coreferent spans of text, it is therefore a somewhat unnatural task. This paper presents an alternative in which we preprocess documents, linking entities to a knowledge base, and turn the coreference annotation task -- in our case limited to pronouns -- into an annotation task where annotators are asked to assign pronouns to entities. Model-based annotation is shown to lead to faster annotation and higher inter-annotator agreement, and we argue that it also opens up for an alternative approach to coreference resolution. We present two new coreference benchmark datasets, for English Wikipedia and English teacher-student dialogues, and evaluate state-of-the-art coreference resolvers on them.  \\ 
\newline  }

\begin{document}

\maketitleabstract

\section{Introduction}

Language comprehension is often seen as the incremental update of a mental model of the situation described in the text \cite{mental-models}. The model is incrementally updated to represent the contents of the linguistic input processed so far, word-by-word or sentence-by-sentence. In this paper, we restrict ourselves to one central feature shared by most theories of mental models: they include a list of entities previously introduced in the text. This corresponds to the {\em constants} of first-order models or the referents associated with different {\em roles} in frame semantics. By models we thus simply mean a set of entities. Obviously, this is not sufficient to represent the meaning of texts, but focusing exclusively on annotating nominal coreference, we can ignore relations and predicates for this work. We will use the term {\em model-based annotation}~to refer to linguistic annotation using model representations to bias or ease the work of the annotators. 

Mental models have previously been discussed in linguistics literature on coreference \cite{tanenhaus-eye-track}. The motivation has often been that some pronouns refer to entities that are not explicitly mentioned in the previous text, but are supposedly available in the reader's mental model of the text, by inference. Consider, for example:

\begin{itemize}
    \item[(1)] I knocked on the door of room 624. {\em He}~wasn't in. 
    \end{itemize}
    
\begin{figure}
    \centering
    \includegraphics[width=\columnwidth]{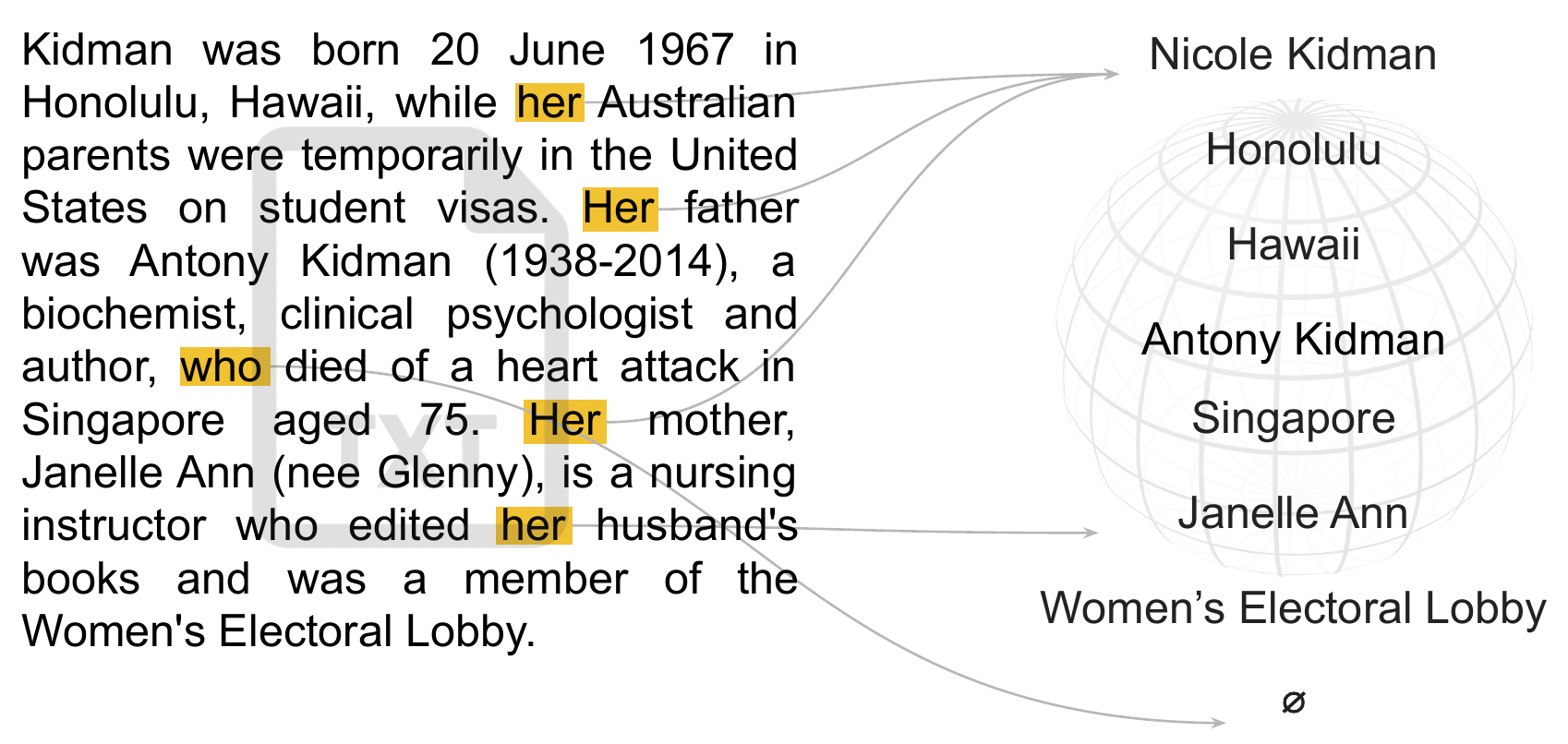}
    \caption{Example of an annotation from the dataset.}
    \label{fig:task}
\end{figure}

The introduction of the referent of {\em he} in (1) is implied by the introduction of the entity {\em room 624}. In this paper, we present a new approach to annotating coreference that enables simple annotation of examples such as (1): Instead of asking an annotator to relate pronouns and previous spans of text, we ask the annotator to link pronouns and entities in document models. Moreover, we argue that model-based annotation reduces the cognitive load of annotators, which we experimentally test by comparing inter-annotator agreement and annotator efficiency across comparable annotation experiments. Fig. \ref{fig:task} showcases a concrete example from the collected dataset.

\paragraph{Contributions} This paper makes a technical contribution, a conceptual contribution, and introduces a novel corpus annotated with coreference to the NLP community: (a) The technical contribution is a novel annotation methodology, where annotation is mediated through a model representation. We believe similar techniques can be developed for other NLP tasks; see \S6 for discussion. (b) The conceptual contribution is a discussion of the importance of mental models in human language processing, and an argument for explicitly representing this level of representation in NLP models. (c) Our corpus consists of manually annotated sentences from English Wikipedia and QuAC \cite{quac}. In addition to the model-based annotations, we also provide the coreference links obtained in our baseline experiments. 

\section{Related Work}
\subsection{Annotation interfaces} 
The idea of easing the cognitive load of annotators by changing the way data is represented, is at the core of many papers on annotation interfaces. Early tools like MMAX2 \cite{mmax2} provide a clean user interface for annotators by highlighting mentions and connecting entity chains to visualize coreference along with helpful features like inter-annotator agreement checker, corpus querying, etc. Newer tools like WebAnno \cite{Yimam:ea:13,callisto} ease the process of annotation by having support for flexible multi-layer annotations on a single document and also provide project management utilities. APLenty \cite{Nghiem:Ananiadou:18} provides automatic annotations for easing annotator load and also has an active learning component which makes the automatic annotations more accurate over time. 

For relieving annotator load, these tools form clusters of coreference such that the annotator can choose to link a mention to one of these clusters. But this is possible only after the clusters are well-formed i.e. after some amount of annotation has taken place. One advantage of our approach is that we provide representatives for each cluster (the entities in the document) right from the start of the annotation process.

\subsection{Mental models in NLP} 

\newcite{Culotta:ea:07} present a probabilistic first-order logic approach to coreference resolution that implicitly relies on mental models. \newcite{Peng:ea:15} focus on hard Winograd-style coreference problems and formulate coreference resolution as an Integer Linear Programming (ILP) to reason about likely models. \newcite{Finkel:Manning:08} also explore simple ILPs over simple first-order models for improving coreference resolution. They obtain improvements by focusing on enforcing transitivity of coreference links. In general, the use of first order models has a long history in NLP, rooted in formal semantics, going back to Fregean semantics. \newcite{Blackburn:Bos:05}, for example, present a comprehensive framework for solving NLP problems by building up first order models of discourses. 

\subsection{Coreference datasets} 
The main resource for English coreference resolution, also used in the CoNLL 2012 Shared Task, is OntoNotes \cite{conll2012}. OntoNotes consists of data from multiple domains, ranging from newswire to broadcast conversations, and also contains annotations for Arabic and Chinese. WikiCoref \cite{wikicoref} is a smaller resource with annotated sentences sampled from English Wikipedia. Our dataset includes paragraphs from all pages annotated in WikiCoref, for comparability with this annotation project. See \S5 for discussion. These are the datasets used below, but alternatives exist: GAP \cite{gap} is another evaluation benchmark, also sampled from Wikipedia and focuses on addressing gender bias in coreference systems. Phrase Detectives \cite{phrase-det} gamifies the creation of anaphoric resources for Wikipedia pages, fiction and art history texts. \newcite{craft} annotate journal articles to create the CRAFT dataset which has structural, coreference and concept annotations. The annotation process of this dataset is similar in spirit to ours as their concept annotations link text mentions to curated ontologies of concepts and entities.

\section {Data collection}
We collect 200 documents\footnote{We use the term {\em document}~to denote a datapoint in our dataset.} from two sources: (i) the summary paragraphs of 100 English Wikipedia documents (30 titles from WikiCoref and 70 chosen randomly), and (ii) the first 100 datapoints from the Question-Answering in Context (QuAC) dataset. Every QuAC document contains a Wikipedia paragraph and QA pairs created by two annotators posing as a student asking questions and a teacher answering the questions by providing short excerpts from the text. Thus the domain of all the documents is English Wikipedia.

\subsection{Design Decisions}
Some Wikipedia articles have short summaries with very few pronouns and some do not have summaries at all. Therefore, for each document chosen randomly, we first verify if it has a summary that contains at least five pronouns. If it does not, we choose another document and repeat this process till we get the required number of documents. We then extract all the entities from every document by parsing URL links present in the document which link to other Wikipedia pages or Wikidata entities. For QuAC documents, where all links are scrubbed, we parse their original Wikipedia pages to get the entities. Lastly we remove all markups, references and lists from the documents. 

We collect a comprehensive list of English pronouns for linking. Some pronouns by their definition, almost never refer to entities. For example, (i) interrogative pronouns: `what', `which', etc., (ii) relative pronouns: `as', `who', etc., and (iii) indefinite pronouns: `anyone', `many', etc. For completeness, we do not remove these words from the list. We however allow the annotators to mark them specifically as \textit{No Reference}.

\subsection{Annotation}
\label{sec:ann}
To test our hypothesis that model-based coreference annotations are faster to create and more coherent, we pose two tasks on Amazon Mechanical Turk (AMT): (i) \textit{Grounded task}: where all the parsed entities from a document are displayed to the annotator for linking with the pronouns, (ii) \textit{Span annotation task}: where the entities are not shown and the annotator is free to choose any span as the antecedent. 30 documents from each source are doubly annotated to compute the inter-annotator agreement and the other 70 were singly annotated.

\begin{figure*}
    \centering
    \includegraphics[width=\textwidth]{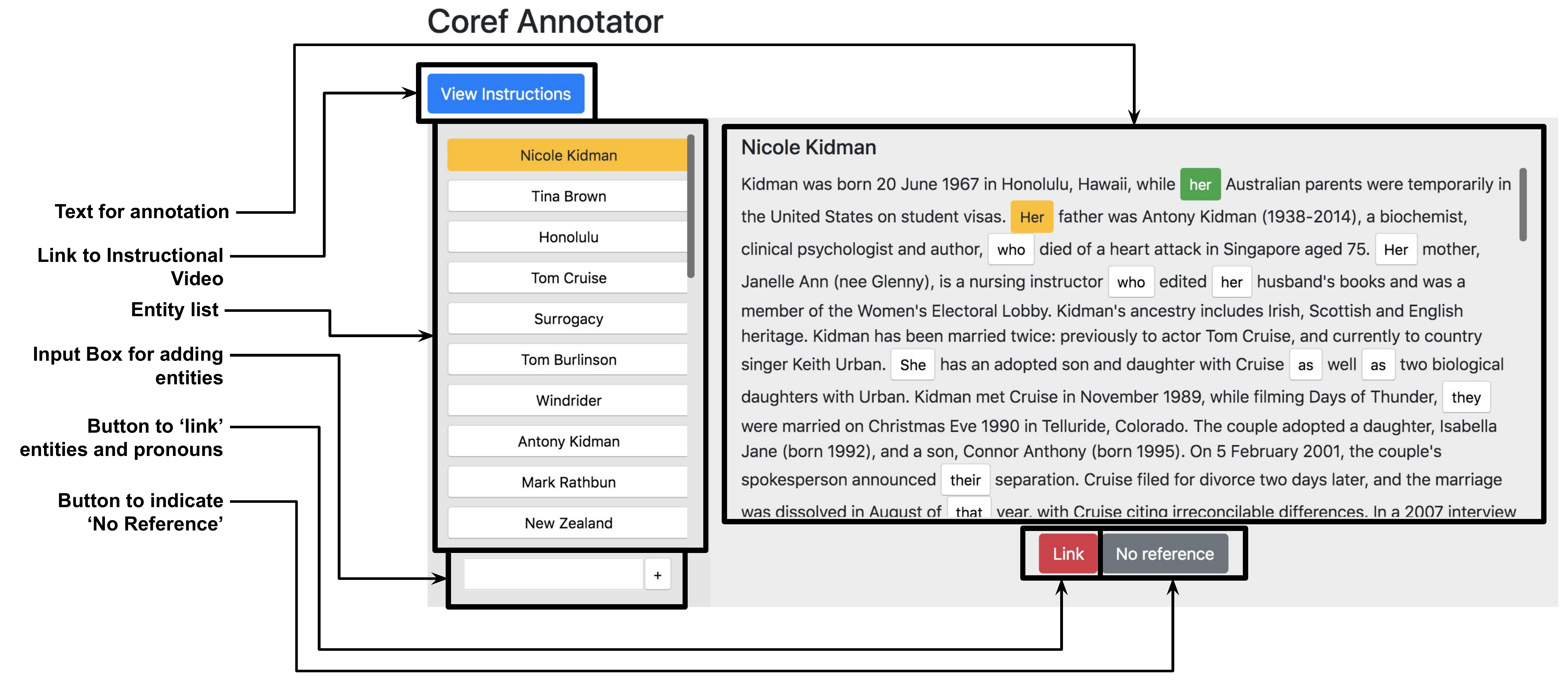}
    \caption{Screen grab of the interface for the grounded-annotation task}
    \label{fig:fixed}
\end{figure*}

\begin{figure*}
    \centering
    \includegraphics[width=\textwidth]{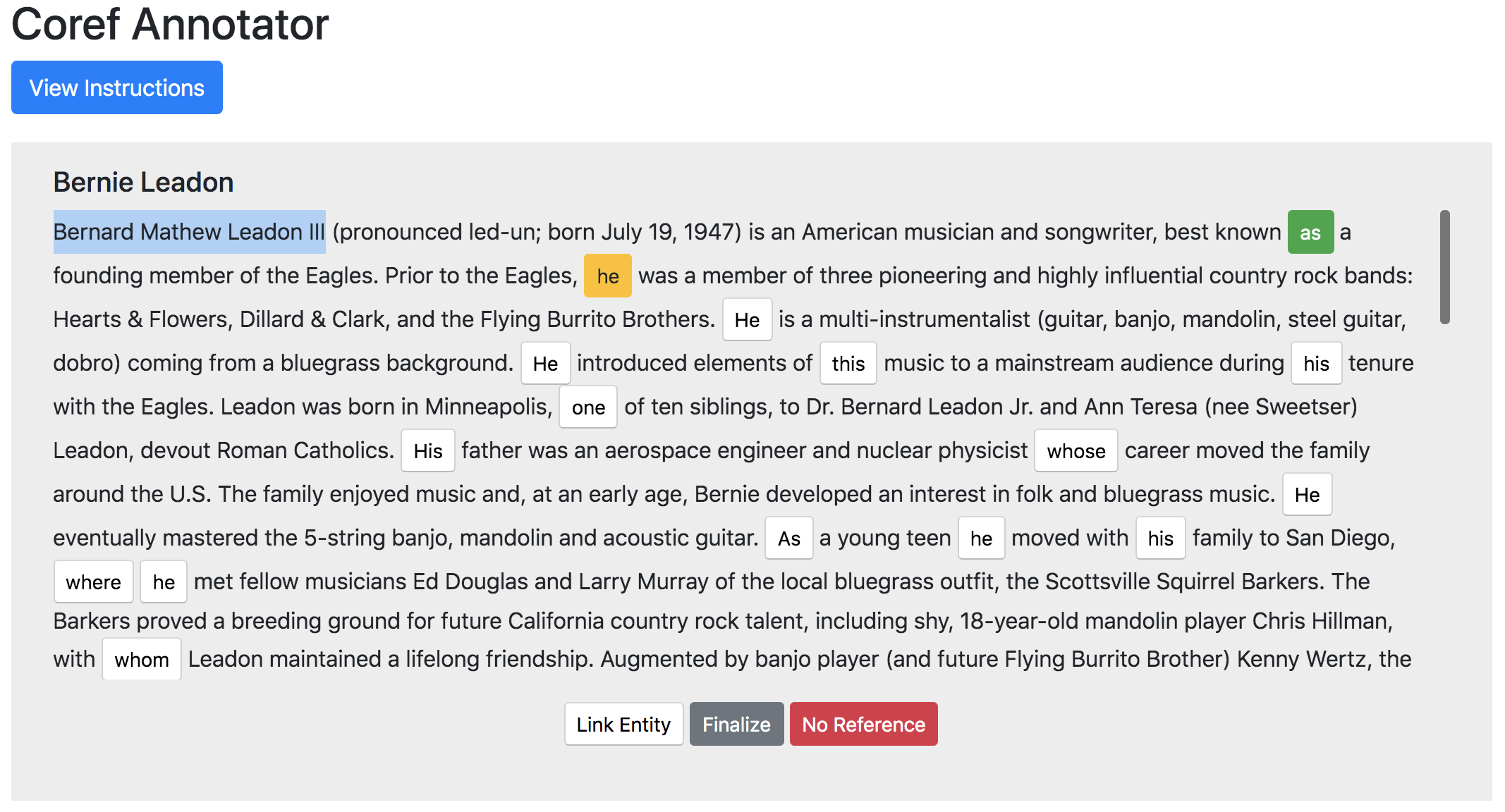}
    \caption{Screen grab of the interface for the span-annotation task}
    \label{fig:select}
\end{figure*}

An annotation tool with two interfaces is built, one for each task, with slight differences between them as shown in Figures \ref{fig:fixed} and \ref{fig:select} respectively. The tool takes in a pre-defined list of mentions (pronouns in our case) which are markable. The annotators can link only these words with coreferent entities. This reduces the cognitive load on the annotators. The annotation process for the two tasks is briefly described below.

\subsubsection{Grounded task}
For this task, the interface (Fig. \ref{fig:fixed}) is split into two parts. A larger part on the right contains the document text and the mention pronouns are highlighted in white. A sidebar on the left is populated with all the entities extracted from the document. In case of missing entities, the annotator has the option of adding one using the input box present at the bottom-left of the screen. The annotators are asked to link the mention pronouns in the document with one or more entities by: (i) clicking on a mention, (ii) clicking on one or more entities; , and (iii) clicking on the red \textit{Link} button. If any mention does not have an antecedent, the annotators are asked to mark them with the grey \textit{No reference} button. The color of the currently selected mention and entities are changed to yellow for convenience. Mentions which are already annotated are marked in green.

\subsubsection{Span annotation task}
In this task, the interface does not have the sidebar (Fig. \ref{fig:select}) and the annotators are free to mark one or more spans in the document as the antecedent(s) for a mention pronoun by selecting the span(s) with their pointers. In a scenario where one mention pronoun has to be linked with multiple antecedents, the annotators have to highlight the spans and click on the white \textit{Link Entity} button multiple times. Therefore, an additional red \textit{Finalize} button is provided to mark the end of one linking episode. Apart from the lack of the entity sidebar and inclusion of the previously mentioned \textit{Finalize} button, all other features of the interface remain the same as those for the Grounded task.

\subsubsection{AMT Details}
The annotation tasks were open only to native English speakers whose approval rate was above 90\%and they had ten minutes to annotate a document. Every fifth document annotated by an annotator was a secret test document for which annotations were known. The annotators were allowed to continue only if there was more than 90\% match between the gold and their annotations. Each task was published 15 days apart to diversify the annotator pool.

\section{Experiments}
\subsection{Inter-annotator agreement} 
As mentioned in Section \ref{sec:ann}, we doubly annotate 30 documents from each source to measure the inter-annotator agreement and the results are presented in Table \ref{tab:iaa}. The numbers clearly indicate that the grounded tasks introduce less uncertainty about the antecedents and hence result in more agreements between the annotators. Ideally the exact match and F$_1$ scores for grounded tasks should be identical. However, the slight difference observed is because of mentions being linked to different similar looking entities. For example, in the sentence ``Harry Potter is a global phenomenon. \textit{It} has captured the imagination of \ldots", the mention \textit{It} can be linked either to Harry Potter -- the movies or Harry Potter -- the books.

\subsection{Annotation times} 
\begin{figure}
    \centering
    \includegraphics[width=\columnwidth]{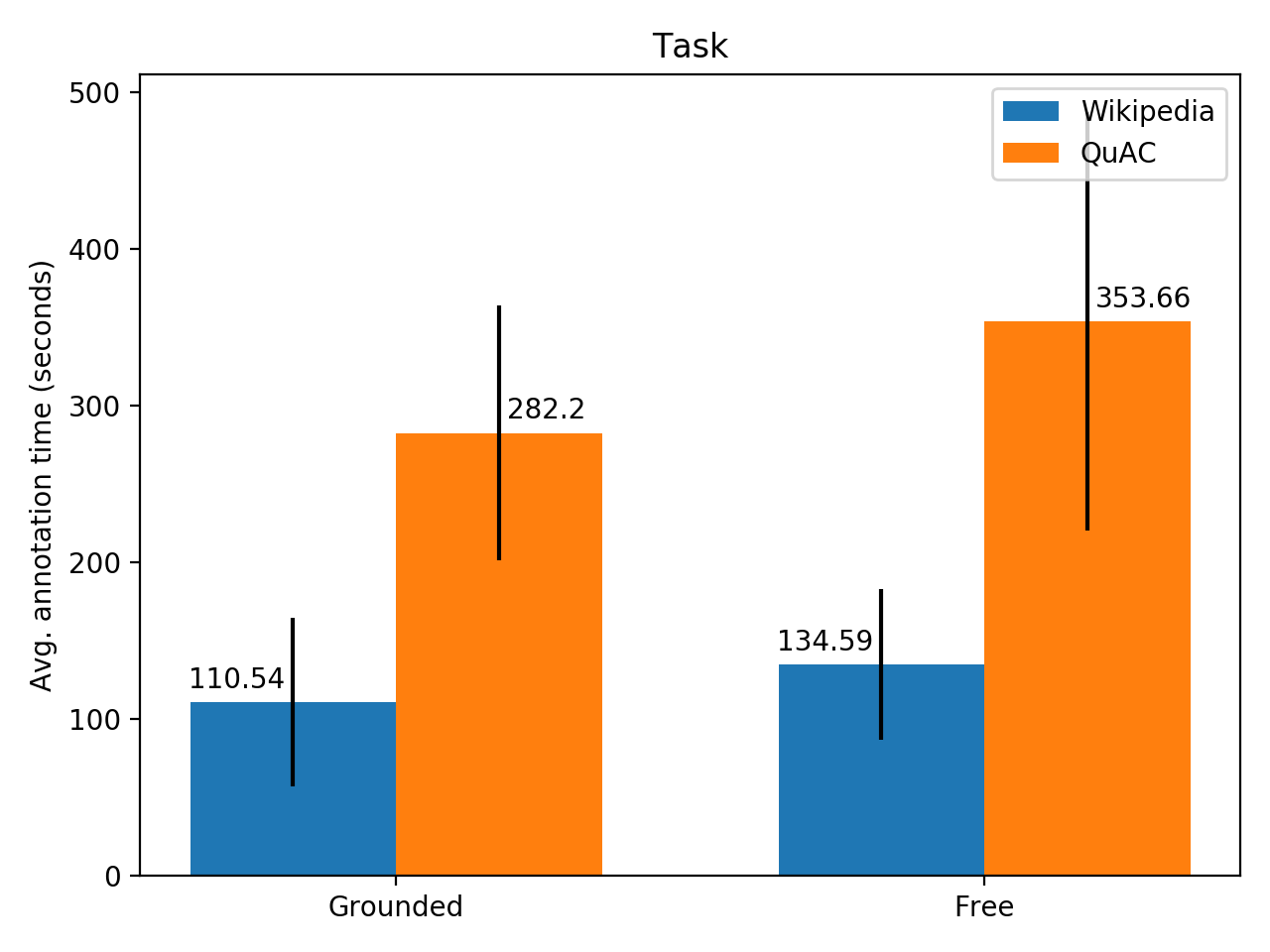}
    \caption{Average annotation times for the two tasks and settings}
    \label{fig:avg_times}
\end{figure}

We can estimate the cognitive load on the annotators by measuring the time taken for marking the documents. Figure \ref{fig:avg_times} shows the mean annotation times and their standard deviations for annotating documents in different settings. In general, QuAC documents require more time and effort to annotate due to the presence of QA pairs which require the annotators to possibly re-read a portion of the context paragraph. Also, it is clear that grounding the document eases the load on annotators irrespective of the source of documents.

\begin{table}[t]
    \centering
    \begin{tabular}{rcc}
        \toprule
            & Exact Match & F$_1$ Score \\
            \midrule
            Wiki grounded & 0.70 & 0.74 \\
            Wiki free & 0.50 & 0.65 \\
            \midrule
            QuAC grounded & 0.65 & 0.67 \\
            QuAC free & 0.52 & 0.64 \\
        \bottomrule
    \end{tabular}
    \caption{Inter-annotator agreement scores}
    \label{tab:iaa}
\end{table}

\subsection{State-of-the-art} 
We run our data through three state-of-the-art coreference resolution systems and report the average precision, recall and F$_1$ scores of three standard metrics: MUC, B$^3$ and CEAFe \cite{coref-metrics}, in Table \ref{tab:results}.\footnote{Converting our grounded data to the OntoNotes format is in some cases lossy, since entity aliases may not perfectly match previous mentions.} While \newcite{clark2016}\footnote{We use an improved implementation available at \href{https://github.com/huggingface/neuralcoref}{https://github.com/huggingface/neuralcoref}.} and \newcite{lee2018} train on OntoNotes 5 to perform both mention detection and entity linking, \newcite{elcoref-qa} use a multi-task architecture for resolving coreference and ellipsis posed as reading comprehension, which is also trained on OntoNotes 5, but uses gold bracketing of the mentions and performs only entity linking.\footnote{This explains the comparatively higher numbers. See discussion in their paper for more details.} The results show that the dataset is hard even for the current state-of-the-art and thus a good resource to evaluate new research.

\begin{table*}[t]
    \centering
    \begin{tabular}{rcccccc}
        \toprule
        \multirow{2}{*}[0pt]{\bf System} & \multicolumn{3}{c}{Wiki} & \multicolumn{3}{c}{QuAC} \\
        \cmidrule{2-7}
        & {\bf P} & {\bf R} & {\bf F$_1$} & {\bf P} & {\bf R} & {\bf F$_1$} \\
        \midrule
        \newcite{clark2016} & 24.72 & 32.87 & 27.95 & 20.15 & 27.98 & 23.39 \\
        \newcite{lee2018} & 21.38 & 37.90 & 26.67 & 17.42 & 39.07 & 23.79 \\
        \newcite{elcoref-qa}$^\ast$ & 43.88 & 48.58 & 45.96 & 46.18 & 46.23 & 46.14 \\
        \bottomrule
    \end{tabular}
    \caption{The macro-averages of MUC, B$^3$, and CEAF$_{\phi_4}$. ($^\ast$assumes gold brackets for mentions.)}
    \label{tab:results}
\end{table*}

\section{Discussion}
The main purpose of this work is to study how humans annotate coreference with and without grounding. Therefore we give freedom to the annotators by asking them to abide by a minimal set of rules. We see interesting annotation patterns in our dataset: 
Generally, the indefinite pronoun `all' is marked as having `No Reference'. But for the sentence ``\ldots Harry Potter, and his friends Hermione Granger and Ron Weasley, \textit{all} of whom \ldots", for example, the pronoun `all' is linked as follows: (i) in the grounded task, the word is linked to three entities -- Harry Potter, Hermione Granger and Ron Weasley, whereas (ii) in the span annotation task, the word is linked to the phrase ``Harry Potter, and his friends Hermione Granger and Ron Weasley". We see that the annotation for the grounded task is cleaner than that for the span annotation task. This effect is observed throughout the dataset. Also, in span annotation tasks, while some annotators link mention pronouns to the first occurrence of an entity, some link them to the latest occurrence, sometimes resulting in multiple clusters instead of one. By design, this is not the case in the grounded tasks.

\subsection{Comparison with WikiCoref}
WikiCoref has 30 annotated pages from English Wikipedia. Our dataset contains 200 documents of which 30 titles are the same as those of WikiCoref. WikiCoref uses the full Wikipedia page for annotation, whereas we extract only the summary paragraphs from each page. WikiCoref doubly annotates only 3 documents for reporting inter-annotator agreement, whereas we do it for 30 documents. The inter-annotator agreements themselves are not comparable because they only report the Kappa coefficient for mention identification which does not occur in our tasks.

\subsection{Generalization to other NLP tasks}
Our first annotation experiments have been limited to coreference for pronouns, but obviously the same technique can be used to annotate other linguistic phenomena involving relations between noun phrases, e.g., other forms of coreference, nominal ellipsis, implicit arguments, or roles of semantic frames. Our models only include individuals or constants, but if we extend our models to also include propositions holding for individuals or between individuals, we could potentially also do grounded annotation of complex verbal phenomena such as VP ellipsis, gapping, sluicing, etc.


\section{Conclusion}
We propose a new way of annotating coreference by grounding the input text to reduce the cognitive load of the annotator. We do this by making the annotators choose the antecedent for mentions from a pre-populated entity list rather than having to select a span manually. We empirically show that annotations performed in this manner are faster and more coherent with higher inter-annotator agreements. We benchmark the collected data on state-of-the-art models and release it in the open domain at \href{https://github.com/rahular/model-based-coref}{https://github.com/rahular/model-based-coref}.

\section*{Acknowledgement} We would like to thank all the members of CoAStaL NLP, Matt Lamm and Vidhyashree Murthy for their valuable feedback on the annotation interface. This work was supported by a Google Focused Research Award. 

\section*{Bibliographical References}
\label{main:ref}

\bibliographystyle{lrec}
\bibliography{ref}

\end{document}